\pdfoutput=1
\documentclass[11pt]{article}

\usepackage[]{acl}

\usepackage{times}
\usepackage{latexsym}
\usepackage{tabularx} 
\usepackage{amsmath}
\usepackage{amsthm}
\usepackage{booktabs}
\usepackage{multirow}
\usepackage{graphicx}
\usepackage{threeparttable}

\usepackage{enumitem}
\usepackage{float}
\usepackage{amsfonts}
\usepackage[ruled,vlined]{algorithm2e}
\usepackage{algorithmic}
\usepackage{url}

\usepackage[T1]{fontenc}
\usepackage[utf8]{inputenc}

\usepackage{microtype}

\usepackage{inconsolata}

\usepackage{graphicx}

\title{
Query Optimization for Parametric Knowledge Refinement in Retrieval-Augmented Large Language Models
}

\author{Youan Cong \ \  
        Pritom Saha Akash \ \ 
        Cheng Wang \ \
        Kevin Chen-Chuan Chang\\
  University of Illinois Urbana-Champaign, USA\\
  \texttt{\{youanc2, pakash2, chengw4, kcchang\}@illinois.edu}}

\begin{document}
\maketitle

\begin{abstract}
We introduce the \textit{Extract-Refine-Retrieve-Read} (ERRR) framework, a novel approach designed to bridge the pre-retrieval information gap in Retrieval-Augmented Generation (RAG) systems through query optimization tailored to meet the specific knowledge requirements of Large Language Models (LLMs). Unlike conventional query optimization techniques used in RAG, the ERRR framework begins by extracting parametric knowledge from LLMs, followed by using a specialized query optimizer for refining these queries. This process ensures the retrieval of only the most pertinent information essential for generating accurate responses. Moreover, to enhance flexibility and reduce computational costs, we propose a trainable scheme for our pipeline that utilizes a smaller, tunable model as the query optimizer, which is refined through knowledge distillation from a larger teacher model. Our evaluations on various question-answering (QA) datasets and with different retrieval systems show that ERRR consistently outperforms existing baselines, proving to be a versatile and cost-effective module for improving the utility and accuracy of RAG systems.
\end{abstract}
\section{Introduction}
The field of natural language processing (NLP) has witnessed transformative advancements in recent years, largely driven by the advent of Large Language Models (LLMs). These models, trained on vast corpora, have demonstrated exceptional capabilities in understanding human text and generating high-quality responses \cite{kaplan2020scaling, clark2022unified}. They have also proven practical and scalable for various downstream NLP tasks, such as conversational response generation, text summarization, and content recommendation, even in few-shot or zero-shot settings \cite{wu2023survey}. Despite their strengths, a key limitation of LLMs lies in their reliance on static training data, which causes them to struggle with dynamic or less commonly known information outside their initial training scope. This limitation often leads to outdated, inaccurate, or entirely fabricated responses—a phenomenon commonly referred to as “hallucination” \cite{lee2018hallucinations}.

To address this issue, Retrieval-Augmented Generation (RAG) \cite{RAG} has emerged as a promising approach to enhance the functionality and reliability of LLMs. By integrating external knowledge sources through retrieval systems, RAG enables LLMs to augment user queries with relevant, up-to-date information. This augmentation allows LLMs to generate more contextually accurate and relevant responses. For instance, in a conversational setting where a user queries an LLM like ChatGPT \cite{gpt} for the latest news, RAG retrieves pertinent articles to supplement the static pre-trained knowledge of the model, thereby mitigating the information gap.
 
While retrieval augmentation has proven effective in mitigating hallucinations, it introduces its own set of challenges. A prominent issue in Retrieval-Augmented Generation (RAG) systems is the \textbf{pre-retrieval gap}—a mismatch between the information retrieved using the original user query and the specific knowledge required to generate optimal responses \cite{ragsurvey}. For instance, consider a document collection containing three passages, labeled Passage \(A\), \(B\), and \(C\), each containing unique knowledge components \(x\), \(y\), and \(z\), respectively. Although all three passages include keywords associated with Knowledge \(z\)—the user’s intended target—a poorly formulated query may lead to retrieving Passage \(A\) or \(B\) instead of the ideal Passage \(C\). This misalignment restricts the LLM reader's ability to generate accurate responses, making the pre-retrieval gap a critical barrier to achieving optimal text generation in RAG systems.

To bridge the pre-retrieval gap, the \textit{Rewrite-Retrieve-Read} (RRR) framework \cite{rrr} introduced query rewriting as a mechanism to optimize user queries and improve their alignment with retrieval systems. However, RRR and similar methods \cite{stepback, ragsurvey} primarily focus on rephrasing or broadening queries, which helps expand the search scope but fails to address the specific knowledge requirements of the LLM reader. Additionally, recent self-prompting techniques \cite{li2022self,wang2023self} have explored using chain-of-thought (CoT) prompts and pseudo-QA pairs to enhance LLM reasoning capabilities by eliciting internal parametric knowledge. While these approaches effectively improve the internal reasoning and explanation capabilities of LLMs for tasks like multi-hop reasoning and open-domain QA, they lack mechanisms for aligning external retrieval queries with the LLM’s knowledge gaps, making them insufficient for resolving the pre-retrieval gap in Retrieval-Augmented Generation (RAG) systems.

To this end, we propose \textit{Extract-Refine-Retrieve-Read} (ERRR), a straightforward yet effective framework designed for retrieval augmentation systems. The ERRR framework is crafted to bridge the pre-retrieval information gap through tailored query optimization and aims to resolve the inherent limitations of RRR by enabling retrieval based on the specific information needs of the LLM reader. Specifically, it initiates by extracting parametric knowledge from LLMs and employs a specialized query optimizer that refines user queries. This refinement either complements or validates the extracted parametric knowledge, ensuring that only essential information is retrieved for generating accurate responses, and minimizing the retrieval of extraneous information that could degrade output quality. 

In addition to its innovative query optimization process, ERRR introduces a trainable scheme to enhance efficiency and adaptability. Recognizing the constraints posed by black-box systems like ChatGPT \cite{gpt}, which are accessible only through inference APIs, ERRR incorporates a smaller, tunable language model as the query optimizer. This trainable component reduces computational costs while offering greater flexibility to customize the retrieval process for diverse queries and knowledge sources. By combining precision in addressing pre-retrieval gaps with cost-effective adaptability, ERRR provides a robust solution for improving retrieval augmentation in LLM-driven systems.

We evaluate ERRR on multiple question-answering (QA) datasets, including HotpotQA \cite{hotpotqa}, AmbigNQ \cite{ambigqa}, and PopQA \cite{popqa}, using both web-based (e.g., Brave Search Engine) and local retrieval systems (e.g., Dense Passage Retrieval \cite{DPR}). Across all tested datasets and retrieval configurations, ERRR consistently outperforms baseline frameworks, such as RRR, in terms of retrieval accuracy and response quality. These results highlight ERRR’s versatility and effectiveness in diverse settings.

In summary, our key contributions are as follows:
(i) We propose \textit{Extract-Refine-Retrieve-Read} (ERRR), a novel framework that optimizes queries to bridge the pre-retrieval gap and enhance RAG systems.
(ii) We demonstrate ERRR’s adaptability across different datasets, retrieval systems, and settings, establishing its robustness and versatility.
(iii) We introduce a trainable ERRR scheme that reduces computational costs while maintaining high performance, making it suitable for real-world applications.
\section{Related work}  
\subsection{Retrieval-Augmented Generation}
The integration of retrieval modules to access relevant contextual knowledge has played a crucial role in enhancing Large Language Models (LLMs) in recent years.  Initially designed for early sequence-to-sequence models, the Retrieval-Augmented Generation (RAG) framework proposed by Lewis et al. \cite{RAG} has gained substantial traction in the era of LLMs. This approach has diversified into a broad array of methods, with ongoing efforts aimed at further augmenting its capabilities. Earlier exploration primarily focused on improving key components, such as upgrading to more powerful pre-trained language models like BERT \cite{bert} as readers or employing advanced dense retrievers for retrieval tasks \cite{DPR}. These retrievers encode documents and inputs into dense vectors, facilitating retrieval based on the similarity between the input and retrieved passages.

Recent studies have shifted focus beyond merely enhancing the retriever or reader components, emphasizing the refinement of pre-retrieval and post-retrieval processes. To address the pre-retrieval gap—the disparity between the information retrievable from original queries and the knowledge required for optimal responses—GenRead \cite{genread} replaces the retrieval module with a knowledgeable LLM, thereby narrowing the gap between the user query and retrieval process. It prompts the LLM to generate contextual information for the query, using these generated documents as retrieval results to formulate the final answer. Self-ask \cite{selfask} proposes an iterative approach using chain-of-thought prompting to generate self-posed questions that refine the response. For the post-retrieval gap—the challenge of creating optimal responses from given information—strategies include document re-ranking or summarization. For instance, PRCA \cite{prca} trains a contextual adapter module to summarize retrieved documents with a black-box LLM reader.

Several studies have also proposed significant modifications to the original RAG pipeline, introducing complex systems that include both pre-retrieval and post-retrieval modules \cite{ragfusion}, and adapting the pipeline into iterative or recursive frameworks \cite{react, selfrag}. While these advanced systems demonstrate notable performance enhancements, they incur substantial costs and typically require multiple interactions with LLMs. In contrast, our work focuses on refining the single-turn RAG framework, introducing a flexible and trainable module adaptable to existing systems.

\subsection{Query Optimization for Retrieval Augmentation}
\begin{figure*}
    \centering
    \includegraphics[width=0.9\linewidth]{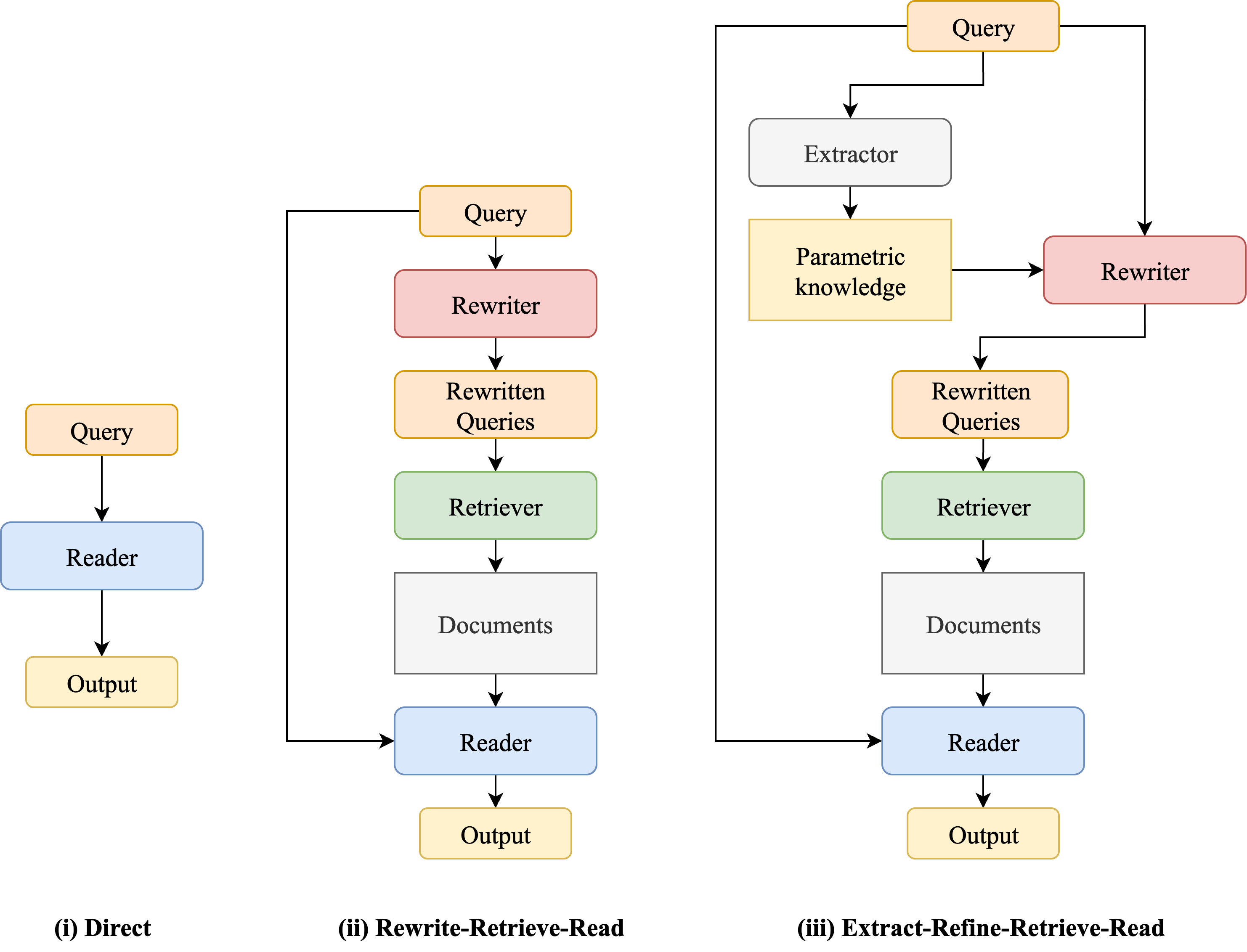}
    \caption{Overview of \textit{Extract-Refine-Retrieve-Read} (ERRR). ERRR leverages parametric knowledge of LLMs and utilizes a specialized query optimizer to retrieve the knowledge that better aligns with the LLM's needs.}
    \label{fig:enter-label}
    \vspace{-5mm}
\end{figure*}

Recent research highlights a significant discrepancy between input queries and LLM readers for RAG systems, especially under the current trend of using off-the-shelf web search tools or black-box LLMs that are difficult to customize \cite{rrr}. Typically, these input queries originate directly from users or specific datasets, which could be either poorly formulated or adhere to a static query format. To overcome these challenges, an effective approach is to optimize the query in the pre-retrieval phase, thereby improving the quality of retrieved information and response generation. The \textit{Rewrite-Retrieve-Read} (RRR) framework, for instance, trains a query rewriting module using an LLM to better align retrieval queries with LLM readers \cite{rrr} that generate the final response, as illustrated in Figure \ref{fig:enter-label}. Additionally, RRR introduces a trainable scheme that employs reinforcement learning with Proximal Policy Optimization to fine-tune a small open-source model based on feedback from the LLM reader, achieving improved results. HyDE addresses the demand for accurate information retrieval by creating hypothetical documents and encoding them through unsupervised contrastive learning for efficient retrieval operations \cite{hyde}. Furthermore, Step-Back Prompting \cite{stepback} converts original queries into high-level abstract questions, aiding LLMs in generating better responses for complex queries requiring abstract thinking.

While these efforts have markedly improved the performance of original RAG systems by focusing on query optimization, they often overlook the importance of synchronizing queries with the specific knowledge requirements of the LLM reader. Unlike the RRR framework, our approach includes an additional parametric knowledge extraction step to assess the knowledge possessed by the LLM. We then perform retrieval based on optimized queries to refine this parametric knowledge, thereby further enhancing retrieval-augmented LLMs.
\section{Methodology}
In this section, we elaborate on the details of \textit{Extract-Refine-Retrieve-Read} (ERRR), a framework for improving retrieval-augmented LLMs through query optimization for parametric knowledge refinement. Section \ref{gap} formally defines the central task addressed by ERRR and introduces its key concepts. The design of the framework is discussed in Section \ref{ERRR}, where we outline a frozen scheme using a black-box LLM reader and standard web search tools. Additionally, Section \ref{Trainable scheme} discusses a trainable scheme of the framework.

\subsection{Pre-retrieval Information Gap} \label{gap}

A task with retrieval augmentation can be formulated as follows. Given an input query $q$, a set of theoretical golden documents $D$ that has the accurate information to answer query $q$, and a ground-truth answer $a$, we denote:
\begin{equation} \label{eq:1}
    \text{LLM}(D, q~|~ \theta) = a
\end{equation}
where $\text{LLM}$ denotes an LLM reader and $\theta$ denotes the parametric knowledge of the LLM. 

However, to obtain the document set $D$, practical implementations often employ a retrieval function $R$ which retrieves documents $R(q)$ from an external knowledge base, and thus the output of a retrieval-augmented system is:
\begin{equation} \label{eq:2}
\text{LLM}(R(q), q ~|~ \theta)
\end{equation}

An inherent challenge arises due to the difference in the quality and relevance of documents retrieved by $R$ compared to the ideal documents set $D$:
\begin{equation} \label{eq:3}
    \text{LLM}(R(q), q ~|~ \theta) \neq \text{LLM}(D, q ~|~ \theta)
\end{equation}

The limitation discussed above describes the problem of the pre-retrieval gap in the original RAG pipeline, wherein the set $R(q)$ may not adequately represent the information necessary for generating the true answer $a$. Therefore, the main objective is to develop a query optimization function $f$ that transforms the initial user query $q$ into one or more optimized queries $f(q)$ such that $R(f(q))$ better approximates the ideal document set $D$. 

Previous work like RRR \cite{rrr} has demonstrated the effectiveness of such query optimization functions, albeit without considering the influence of $\theta$. To this end, ERRR introduces a more tailored query optimization function $f'$ that utilizes the parametric knowledge $\theta$ to perform the query optimization and retrieve external knowledge that refines $\theta$ and better aligns with its needs. This can be formulated as:
\begin{equation} \label{eq:3}
    \text{LLM}(R(f'(C, q)), q ~|~ \theta) 
\end{equation}
where $$C = E(q~|~\theta)$$
and $E$ denotes the parametric knowledge extraction function.

\subsection{Extract-Refine-Retrieve-Read}\label{ERRR}

\textit{Extract-Refine-Retrieve-Read} consists of a four-step pipeline: Parametric Knowledge Extraction, Query Optimization for Parametric Knowledge Refinement, Retrieval, and Generation, as depicted in Figure \ref{fig:enter-label}. Detailed technical implementation for each step, covering the models, prompting techniques and training setup, is provided in Section \ref{sec:implementation}.

\noindent \textbf{Parametric Knowledge Extraction}  \quad 
Previous studies such as GenRead \cite{genread} and HyDE \cite{hyde} demonstrate that LLMs may possess substantial parametric knowledge capable of addressing user inquiries, particularly on popular topics. Inspired by the prompting methods outlined in GenRead, our approach involves a direct strategy where we prompt the LLM reader to produce a pseudo-contextual document containing all the background information. We consider these pseudo-contextual documents as a representation of the LLM's abstracted parametric knowledge. Although these documents may contain inaccuracies, they provide essential contextual information related to the original queries.

\noindent \textbf{Query Optimization}  \quad 
In this step, we employ an LLM as the query optimizer for parametric knowledge refinement. We prompt the query optimizer to produce one or more optimized queries seeking external knowledge that either validates or supplements the existing parametric knowledge, especially focusing on the validation of time-sensitive information.

\noindent \textbf{Retrieval}  \quad
To illustrate the adaptability of our module across various retrieval systems and data sources, we utilize two types of retrievers: a black-box web search tool and a local dense retrieval system, which are then combined with the original query for processing by the LLM reader.

\noindent \textbf{Generation}  \quad
We employ an LLM reader to generate the final answer using both the retrieved documents and the original query. Our prompting strategy involves straightforward instructions followed by 1-3 few-shot examples for question answering. These examples are consistently used within each dataset but vary across different datasets to maintain control over the task-specific output format from the LLM reader—for instance, the responses are expected to be concise in certain QA tasks, usually only one or a few words.

\subsection{Trainable Scheme} \label{Trainable scheme}
Given that many powerful LLMs operate as black-box systems, significant challenges such as high computational costs, customization limitations, copyright issues, and connectivity problems have arisen. To address these issues, alongside the conventional frozen scheme, we propose a trainable scheme for our pipeline. Specifically, we fine-tuned a smaller, trainable model using knowledge distillation from a high-performing teacher LLM. We leveraged the teacher's outputs as a strong learning template, intensively training the student model on a distillation dataset of QA questions and generated responses to learn the intricate nuances of query optimization. This streamlined model is then integrated into our pipeline to fulfill the role of query rewriting, originally handled by a frozen LLM.

\begin{table*}[tbh]
	\centering\small
        \resizebox{\linewidth}{!}{
	\begin{tabularx}{0.9\linewidth}{X}
		\toprule
		\textbf{Direct Prompt}   \\
             \midrule
		 Answer the question in the following format, end the answer with '**'. \{demonstration\} Question: \{$x$\} Answer:
		\\
		\midrule
		\textbf{Reader Prompt for Retrieval Augmentation Generation} \\
            \midrule
            Answer the question in the following format, end the answer with '**'. \{demonstration\} Question: \{$doc$\} \{$x$\} Answer: \\
            \midrule
            \textbf{Prompt for RRR Query Rewriter}
            \\
            \midrule
            Think step by step to answer this question, and provide search engine queries for knowledge that you need. Split the queries with ';' and end the queries with '**'. \{demonstration\} Question: \{$x$\} Answer: 
            \\
            \midrule
            \textbf{Prompt for Parametric Knowledge Extraction} \\
            \midrule
            Generate a background document from web to answer the given question. \{$x$\} \\
            \midrule
            \textbf{Prompt for ERRR Query Optimizer} \\
            \midrule
            Address the following questions based on the contexts provided. Identify any missing information or areas requiring validation, especially if time-sensitive data is involved. Then, formulate several specific search engine queries to acquire or validate the necessary knowledge. Split the queries with ';' and end the queries with '**'. \{demonstration\} Context: \{Parametric Knowledge\} Question: \{$x$\} Queries: 
            \\
		\bottomrule
	\end{tabularx}}
	\caption{List of prompts used.}
        \label{prompt}
        \vspace{-1.2em}
\end{table*}%
\section{Experiments}

\subsection{Datasets and Metrics}
ERRR is assessed on three open-domain question-answering (QA) datasets: AmbigQA \cite{ambigqa}, PopQA \cite{popqa}, and HotpotQA \cite{hotpotqa}. Each dataset serves to test different capabilities of the ERRR framework. (i) The AmbigNQ dataset is the disambiguated variant of the Natural Questions (NQ) dataset, where ambiguous questions from NQ are refined into specific queries with minimal constraints. Consistent with procedures used in RRR, we evaluated ERRR using the first 1000 samples of the test set. (ii) PopQA features simpler questions that focus on less popular knowledge topics compared to other QA tasks. Due to the high similarity in sample distributions, we assessed only the first 997 samples of the test set. (iii) The HotPotQA dataset contains complex questions that require multi-hop reasoning. We conducted evaluations across the entire test set. Following the metric usage for three datasets, our method is evaluated by exact match score $EM$ and $F_1$ score.

\subsection{Baselines and Proposed Frameworks}
We evaluated 7 baselines and proposed frameworks, as detailed below:
(i) \textbf{Direct}: Directly calling GPT-3.5-Turbo to answer questions.
(ii) \textbf{RAG}: The classic Retrieval-Augmented Generation framework \cite{RAG}. The original user queries are used for retrieval and fed directly to the LLM reader to generate output.
(iii) \textbf{ReAct}: A modified RAG framework that intertwines the reasoning and acting capabilities of LLMs to create a more cohesive and effective approach \cite{react}. This framework can iteratively perform reasoning prompts and actions, such as information retrieval, serving as our comparison baseline.
(iv) \textbf{Frozen RRR}:  \textit{Rewrite-Retrieve-Read} framework \cite{rrr} with a frozen configuration. It employs GPT-3.5-Turbo to rewrite the query and retrieve relevant documents based on these rewritten queries. Then the original query and retrieved documents are used for reading. This serves as our baseline for comparison.
(v) \textbf{Trainable RRR}: Trainable \textit{rewrite-retrieve-read} framework, initiating with a supervised fine-tuned T5-large model. It then applies reinforcement learning to better align the retriever and rewriter using Proximal Policy Optimization (PPO). This serves as our baseline for comparison.
(vi) \textbf{Frozen ERRR}: \textit{Extract-Refine-Retrieve-Read} framework with a frozen configuration, as described in Section \ref{ERRR}.
(vii) \textbf{Trainable ERRR}: Trainable \textit{Extract-Refine-Retrieve-Read} framework, as described in Section \ref{Trainable scheme}.

These frameworks are evaluated using a web search tool or a local retriever with a static corpus, as described in Section \ref{ERRR}. Due to resource limitations, some frameworks were not evaluated under the local dense retriever setting.

\begin{table*}
    \centering
    \begin{tabular}{lcccccc}
    \toprule
     {} & \multicolumn{2}{c}{\textbf{AmbigQA}} & \multicolumn{2}{c}{\textbf{PopQA}} & \multicolumn{2}{c}{\textbf{HotPotQA}} \\
     \cmidrule(lr){2-3} \cmidrule(lr){4-5} \cmidrule(lr){6-7}
     \textbf{Methods} & EM & F1 & EM & F1 & EM & F1 \\
     \midrule
     Direct&0.391&0.4996&0.392&0.4289&0.311&0.4178\\
     RAG&0.473&0.5842&0.425&0.4704&0.329&0.4424\\
     ReAct&0.477&0.5787&0.451&0.4917&0.344$^*$&0.4649$^*$\\
     Frozen RRR&0.452&0.5577&0.445&0.4904&0.337&0.4567\\
     Trainable RRR&0.460&0.5777&0.389&0.4238&0.337&0.4548\\
     Frozen ERRR&0.4815&0.5823&0.480&0.5256&0.369&0.4941\\
     Trainable ERRR&\textbf{0.4975}&\textbf{0.5988}&\textbf{0.485}&\textbf{0.5309}&\textbf{0.372}&\textbf{0.4989}\\
    \bottomrule
    \end{tabular}
    \caption{The retrieval system in the above methods is Brave Search API.  "Frozen" indicates the rewriter or the query optimizer is GPT-3.5-Turbo, while "Trainable" refers to the rewriter or the query optimizer is a supervised fine-tuned T5 model. Trainable RRR is also trained using proximal policy optimization (PPO) following the original paper. '*' indicates that it is evaluated on 500 random questions drawn from HotPotQA due to resource limitation.}
    \label{tab:my_label_websearch}
    \vspace{-5mm}
\end{table*}

\subsection{Implementation Details}\label{sec:implementation}

For all baselines, we utilized GPT-3.5-Turbo as the primary LLM and followed the implementation details from their respective original papers. GPT-3.5 Turbo was chosen for its balance of performance and cost, aligning with our focus on optimizing retrieval-augmented generation systems rather than benchmarking generative models themselves. While GPT-4 offers improved capabilities, our emphasis remained on augmenting the model’s utility through query optimization. Notably, for the Trainable RRR, we employed the supervised fine-tuned T5 model checkpoint as the base model. This checkpoint, open-sourced by the original authors, has been warmed up and fine-tuned on multiple datasets to function as the query rewriter. Then we replicated their reinforcement learning process since we replaced the original search tool with the Brave Search Engine. This training was conducted on the first 1000 data points for each dataset evaluated, with The training parameters set as follows: a learning rate of 2e-5, 3 epochs, and a batch size of 8. 

For our proposed methods ERRR, in addition to the settings mentioned in Section \ref{ERRR}, the following sections outline technical details:

\noindent \textbf{Parametric Knowledge Extraction}  \quad
To perform parametric knowledge extraction, we use the same prompts from the GenRead paper and choose the top prompt that is most likely to produce pseudo-contextual documents. We outline these extraction prompts in Table \ref{prompt}.

\noindent \textbf{Query Optimization}  \quad
Our specific prompt structure is detailed in Table \ref{prompt}, where {demonstration} consists of 2 manually crafted examples. These examples are consistently used across all tests and primarily serve as one or few-shot examples for the query optimizer.

\noindent \textbf{Retrieval}  \quad
For our web search engine, we opt for the Brave Search Engine, which, although it may provide slightly lower quality results compared to major competitors like Google or Bing, offers a significantly more cost-effective API. This search API retrieves website snippets, simulating a typical user experience of entering a query in a search engine, pressing Enter, and reviewing the top results at a glance. For local retrieval, we utilize WikiDPR, a specialized subset of Wikipedia collections tailored for the Dense Passage Retrieval (DPR) model \cite{DPR}. This database consists of 21 million passages from Dec. 20, 2018, each limited to 100 words, along with their 768-dimensional embedded vectors. The retrieval process involves converting a query into a DPR embedding and finding the top k vectors with the closest L2 distances. For both systems, we retrieve the top 5 results, concatenate them with the original query, and feed them to the LLM reader.

\noindent \textbf{Generation}  \quad
Although different prompting strategies may influence the performance of the question-answering task, this aspect is not the primary focus of our study, so we adhere to the same answer prompts used in the RRR \cite{rrr} framework. The prompts we used are detailed in Table \ref{prompt}.

\noindent \textbf{Trainable Scheme}  \quad
For Trainable ERRR, we employ T5-Large \cite{t5}, an open-source model with 770 million parameters, as the query optimizer. We fine-tune this student model using knowledge distillation from GPT-3.5-Turbo. The distillation dataset was assembled by selecting questions from training sets of each QA dataset, with GPT-3.5-Turbo generating the responses under identical settings utilized in the frozen scheme. We also devised a short eliciting prompt, "Rewrite better search queries to acquire or validate the knowledge needed for the question:", serving as an instruction prefix to guide T5 to adapt to the task. To ensure optimal task-specific outcomes, separate T5 models were trained with 3 epochs for each QA dataset, with a learning rate of 1e-4 and a batch size of 4.

\begin{table*}
    \centering
    \begin{tabular}{lcccccc}
    \toprule
     {} & \multicolumn{2}{c}{\textbf{AmbigQA}} & \multicolumn{2}{c}{\textbf{PopQA}} & \multicolumn{2}{c}{\textbf{HotPotQA}} \\
     \cmidrule(lr){2-3} \cmidrule(lr){4-5} \cmidrule(lr){6-7}
     \textbf{Methods} & EM & F1 & EM & F1 & EM & F1 \\
     \midrule
     Direct & 0.391 & 0.4996 & 0.392 & 0.4289 & 0.311 & 0.4178 \\
     Frozen RRR & 0.438 & 0.5373 & 0.378 & 0.4517 & 0.289 & 0.3926 \\
     Trainable RRR & 0.414 & 0.5203 & 0.365 & 0.4242 & 0.282 & 0.3764 \\
     Frozen ERRR & 0.448 & 0.5473 & 0.419 & 0.4685 & 0.337 & 0.4482 \\
     Trainable ERRR & \textbf{0.4595} & \textbf{0.5777} & \textbf{0.426} & \textbf{0.4694} & \textbf{0.338} & \textbf{0.4499} \\
    \bottomrule
    \end{tabular}
    \caption{Evaluations with WikiDPR as local retrievers. The other setting is the same as Table \ref{tab:my_label_websearch}. Due to resource limitations, some baselines were not fully evaluated under this setting.}
    \label{tab:my_label_dense}
\end{table*}

\subsection{Result}

\begin{table*}
    \centering
    \begin{tabular}{lcccc}
    \toprule
        {} &Frozen ERRR&Trainable ERRR&ReAct&Self-RAG\\
    \midrule
        Cost&\$0.62&\$0.53&\$1.05&\$1.65\\
        Latency&148s&140s& 202s& 270s\\
    \bottomrule
    \end{tabular}
    \caption{The total cost and total latency of each method that is evaluated on 200 randomly drawn data points from HotPotQA.}
    \label{cost-latency}
    \vspace{-5mm}
\end{table*}

The experimental results across three datasets and two retrieval tools are presented in Table \ref{tab:my_label_websearch} and Table \ref{tab:my_label_dense}. The Frozen ERRR framework consistently outperforms all baseline methods—Direct, Frozen RRR, and Trainable RRR—regardless of the retrieval system used. These results highlight the effectiveness of addressing the pre-retrieval information gap, demonstrating ERRR’s adaptability across diverse retrieval systems and datasets. Furthermore, the Trainable ERRR framework achieves even better performance, surpassing all baselines and its teacher model (GPT-3.5 Turbo) across all three datasets. We attribute this improvement to the distillation process, which enables the student model (fine-tuned T5) to generalize better by focusing on critical features while filtering out irrelevant information. This distilled representation allows the model to adapt more effectively to specific query optimization tasks, potentially compressing and refining the teacher’s insights into a more efficient form.

The impact of the ERRR framework is more pronounced in web search retrieval systems, as evidenced by the greater performance enhancement observed in Table \ref{tab:my_label_websearch} compared to dense retrievers in Table \ref{tab:my_label_dense}. This is likely due to the higher quality and broader knowledge span of web-based retrieval systems compared to the static 2018 Wikipedia corpus used for dense retrieval. Notably, the results show that both Frozen RRR and Trainable RRR underperform the Direct method in the PopQA and HotPotQA datasets when using dense retrieval. This underperformance can be attributed to the low-quality results retrieved from the outdated and limited corpus, which includes only Wikipedia passages of constrained length and scope. These limitations lead to an increased retrieval of irrelevant documents, hindering the LLM from answering questions correctly.

In contrast, ERRR demonstrates resilience under such conditions. By optimizing queries to align with the LLM’s informational needs, ERRR reduces the retrieval of irrelevant passages, mitigating distractions caused by lower-quality retrieval. This robustness is particularly valuable when operating on suboptimal document collections, as it ensures performance gains even in challenging retrieval scenarios. A detailed case study, provided in Appendix \ref{sec:appendix}, further illustrates how ERRR generates precise queries that enhance retrieval effectiveness and improve final answers, even when the retrieved content includes inaccuracies.

\subsection{Cost and Latency}
Given our method's emphasis on a conventional single-turn pipeline, it demonstrates superior performance in terms of cost and latency when compared to certain advanced and iterative RAG frameworks. To underscore the cost-efficiency and flexibility of our approach, we conducted a comparative analysis with ReAct \cite{react} and Self-RAG \cite{selfrag}. This experiment was carried out on 200 randomly selected questions from HotPotQA. The results presented in Table \ref{cost-latency} highlight that while still maintaining commendable performance, Frozen ERRR exhibits significantly lower costs, faster processing times, and greater efficiency than other iterative frameworks. Moreover, Trainable ERRR has the potential to further reduce costs, particularly for large datasets, by leveraging an already fine-tuned query optimizer, thereby eliminating an additional LLM call to GPT-3.5-Turbo.

\section{Conclusion}
In this paper, we present \textit{Extract-Refine-Retrieve-Read} (ERRR) framework for Retrieval-Augmented Generation (RAG) systems. The ERRR framework is designed to optimize queries, aligning them closely with the specific informational needs of large language models (LLMs) to enhance retrieval augmentation effectiveness. Our experimental results demonstrate that our method surpasses both the naive LLM  and native query rewriting framework \textit{Rewrite-Retrieve-Read} on benchmark datasets such as AmbigQA \cite{ambigqa}, PopQA \cite{popqa} and HotPotQA \cite{hotpotqa}, utilizing both web search tools and a dense retriever with local static corpus. These results demonstrated ERRR's remarkable adaptability across a variety of settings, data sources, and retrieval systems. This flexibility ensures that ERRR can be effectively implemented in diverse operational environments, making it a potential and adaptable component for inclusion in more advanced RAG systems. Additionally, we have developed and implemented a trainable scheme for the ERRR framework. This approach is both cost-effective and efficient as it relies on only a fine-tuned T5 model trained on a moderately sized dataset and surpasses the performance of the frozen GPT-3.5-Turbo.

\section{Limitation}

We acknowledge the existence of more sophisticated Retrieval-Augmented Generation (RAG) systems such as Self-RAG \cite{selfrag} and CRAG \cite{CRAG}. These advanced systems typically require iterative invocations of the entire pipeline to refine their answers, resulting in exceptionally high computational demands. Due to these computational constraints, our study focused solely on scenarios that operate in a single-turn manner, wherein each module is invoked only once. 
Additionally, our trainable query optimizer does not employ any reinforcement learning (RL) techniques to further refine its alignment with the reader, a decision driven by resource constraints and observed performance degradation when training on a small subset of the dataset using Proximal Policy Optimization (PPO) \cite{schulman2017proximal}.

The effectiveness of the ERRR framework is also inherently tied to the capabilities of the underlying LLMs used for knowledge extraction, query optimization, and final generation. The quality of the extracted knowledge and the refined queries are dependent on the base model's own internal knowledge and reasoning abilities. Consequently, the framework's overall performance may vary significantly when implemented with different or future LLMs.

These limitations point toward several directions for future work. Beyond addressing the pre-retrieval gap, future research could also tackle post-retrieval challenges by exploring methods to better re-rank, filter, or synthesize retrieved documents before they are passed to the reader. Furthermore, ERRR could be incorporated as a modular component within more advanced, iterative RAG systems to create a more holistic and powerful RAG pipeline. Finally, exploring novel RL algorithms tailored for this framework could lead to further improvements for the trainable optimizer, allowing it to better realize its potential for adaptation.
\section{Acknowledgement}
This material is based upon work supported by the National Science Foundation IIS 16-19302 and IIS 16-33755, Zhejiang University ZJU Research 083650, IBM-Illinois Center for Cognitive Computing Systems Research (C3SR) and IBM-Illinois Discovery Accelerator Institute (IIDAI), grants from eBay and Microsoft Azure, UIUC OVCR CCIL Planning Grant 434S34, UIUC CSBS Small Grant 434C8U, and UIUC New Frontiers Initiative. Any opinions, findings, conclusions, or recommendations expressed in this publication are those of the author(s) and do not necessarily reflect the views of the funding agencies.
\bibliography{anthology}

\begin{thebibliography}{26}
\providecommand{\natexlab}[1]{#1}

\bibitem[{Asai et~al.(2023)Asai, Wu, Wang, Sil, and Hajishirzi}]{selfrag}
Akari Asai, Zeqiu Wu, Yizhong Wang, Avirup Sil, and Hannaneh Hajishirzi. 2023.
\newblock \href {https://arxiv.org/abs/2310.11511} {Self-rag: Learning to retrieve, generate, and critique through self-reflection}.
\newblock \emph{Preprint}, arXiv:2310.11511.

\bibitem[{Clark et~al.(2022)Clark, de~Las~Casas, Guy, Mensch, Paganini, Hoffmann, Damoc, Hechtman, Cai, Borgeaud et~al.}]{clark2022unified}
Aidan Clark, Diego de~Las~Casas, Aurelia Guy, Arthur Mensch, Michela Paganini, Jordan Hoffmann, Bogdan Damoc, Blake Hechtman, Trevor Cai, Sebastian Borgeaud, et~al. 2022.
\newblock Unified scaling laws for routed language models.
\newblock In \emph{International conference on machine learning}, pages 4057--4086. PMLR.

\bibitem[{Devlin et~al.(2019)Devlin, Chang, Lee, and Toutanova}]{bert}
Jacob Devlin, Ming-Wei Chang, Kenton Lee, and Kristina Toutanova. 2019.
\newblock \href {https://doi.org/10.18653/v1/N19-1423} {{BERT}: Pre-training of deep bidirectional transformers for language understanding}.
\newblock In \emph{Proceedings of the 2019 Conference of the North {A}merican Chapter of the Association for Computational Linguistics: Human Language Technologies, Volume 1 (Long and Short Papers)}, pages 4171--4186, Minneapolis, Minnesota. Association for Computational Linguistics.

\bibitem[{Gao et~al.(2023)Gao, Ma, Lin, and Callan}]{hyde}
Luyu Gao, Xueguang Ma, Jimmy Lin, and Jamie Callan. 2023.
\newblock \href {https://doi.org/10.18653/v1/2023.acl-long.99} {Precise zero-shot dense retrieval without relevance labels}.
\newblock In \emph{Proceedings of the 61st Annual Meeting of the Association for Computational Linguistics (Volume 1: Long Papers)}, pages 1762--1777, Toronto, Canada. Association for Computational Linguistics.

\bibitem[{Gao et~al.(2024)Gao, Xiong, Gao, Jia, Pan, Bi, Dai, Sun, Wang, and Wang}]{ragsurvey}
Yunfan Gao, Yun Xiong, Xinyu Gao, Kangxiang Jia, Jinliu Pan, Yuxi Bi, Yi~Dai, Jiawei Sun, Meng Wang, and Haofen Wang. 2024.
\newblock \href {https://arxiv.org/abs/2312.10997} {Retrieval-augmented generation for large language models: A survey}.
\newblock \emph{Preprint}, arXiv:2312.10997.

\bibitem[{Kaplan et~al.(2020)Kaplan, McCandlish, Henighan, Brown, Chess, Child, Gray, Radford, Wu, and Amodei}]{kaplan2020scaling}
Jared Kaplan, Sam McCandlish, Tom Henighan, Tom~B Brown, Benjamin Chess, Rewon Child, Scott Gray, Alec Radford, Jeffrey Wu, and Dario Amodei. 2020.
\newblock Scaling laws for neural language models.
\newblock \emph{arXiv preprint arXiv:2001.08361}.

\bibitem[{Karpukhin et~al.(2020)Karpukhin, Oguz, Min, Lewis, Wu, Edunov, Chen, and Yih}]{DPR}
Vladimir Karpukhin, Barlas Oguz, Sewon Min, Patrick Lewis, Ledell Wu, Sergey Edunov, Danqi Chen, and Wen-tau Yih. 2020.
\newblock \href {https://doi.org/10.18653/v1/2020.emnlp-main.550} {Dense passage retrieval for open-domain question answering}.
\newblock In \emph{Proceedings of the 2020 Conference on Empirical Methods in Natural Language Processing (EMNLP)}, pages 6769--6781, Online. Association for Computational Linguistics.

\bibitem[{Lee et~al.(2018)Lee, Firat, Agarwal, Fannjiang, and Sussillo}]{lee2018hallucinations}
Katherine Lee, Orhan Firat, Ashish Agarwal, Clara Fannjiang, and David Sussillo. 2018.
\newblock Hallucinations in neural machine translation.

\bibitem[{Lewis et~al.(2020)Lewis, Perez, Piktus, Petroni, Karpukhin, Goyal, K\"{u}ttler, Lewis, Yih, Rockt\"{a}schel, Riedel, and Kiela}]{RAG}
Patrick Lewis, Ethan Perez, Aleksandra Piktus, Fabio Petroni, Vladimir Karpukhin, Naman Goyal, Heinrich K\"{u}ttler, Mike Lewis, Wen-tau Yih, Tim Rockt\"{a}schel, Sebastian Riedel, and Douwe Kiela. 2020.
\newblock Retrieval-augmented generation for knowledge-intensive nlp tasks.
\newblock In \emph{Proceedings of the 34th International Conference on Neural Information Processing Systems}, NIPS '20, Red Hook, NY, USA. Curran Associates Inc.

\bibitem[{Li et~al.(2022)Li, Wang, Zhang, and Zhao}]{li2022self}
Junlong Li, Jinyuan Wang, Zhuosheng Zhang, and Hai Zhao. 2022.
\newblock Self-prompting large language models for zero-shot open-domain qa.
\newblock \emph{arXiv preprint arXiv:2212.08635}.

\bibitem[{Ma et~al.(2023)Ma, Gong, He, Zhao, and Duan}]{rrr}
Xinbei Ma, Yeyun Gong, Pengcheng He, Hai Zhao, and Nan Duan. 2023.
\newblock \href {https://doi.org/10.18653/v1/2023.emnlp-main.322} {Query rewriting in retrieval-augmented large language models}.
\newblock In \emph{Proceedings of the 2023 Conference on Empirical Methods in Natural Language Processing}, pages 5303--5315, Singapore. Association for Computational Linguistics.

\bibitem[{Mallen et~al.(2022)Mallen, Asai, Zhong, Das, Khashabi, and Hajishirzi}]{popqa}
Alex Mallen, Akari Asai, Victor Zhong, Rajarshi Das, Daniel Khashabi, and Hannaneh Hajishirzi. 2022.
\newblock When not to trust language models: Investigating effectiveness of parametric and non-parametric memories.
\newblock \emph{arXiv preprint arXiv:2212.10511}.

\bibitem[{Min et~al.(2020)Min, Michael, Hajishirzi, and Zettlemoyer}]{ambigqa}
Sewon Min, Julian Michael, Hannaneh Hajishirzi, and Luke Zettlemoyer. 2020.
\newblock Ambigqa: Answering ambiguous open-domain questions.
\newblock \emph{arXiv preprint arXiv:2004.10645}.

\bibitem[{Ouyang et~al.(2022)Ouyang, Wu, Jiang, Almeida, Wainwright, Mishkin, Zhang, Agarwal, Slama, Ray, Schulman, Hilton, Kelton, Miller, Simens, Askell, Welinder, Christiano, Leike, and Lowe}]{gpt}
Long Ouyang, Jeff Wu, Xu~Jiang, Diogo Almeida, Carroll~L. Wainwright, Pamela Mishkin, Chong Zhang, Sandhini Agarwal, Katarina Slama, Alex Ray, John Schulman, Jacob Hilton, Fraser Kelton, Luke Miller, Maddie Simens, Amanda Askell, Peter Welinder, Paul Christiano, Jan Leike, and Ryan Lowe. 2022.
\newblock \href {https://arxiv.org/abs/2203.02155} {Training language models to follow instructions with human feedback}.
\newblock \emph{Preprint}, arXiv:2203.02155.

\bibitem[{Press et~al.(2023)Press, Zhang, Min, Schmidt, Smith, and Lewis}]{selfask}
Ofir Press, Muru Zhang, Sewon Min, Ludwig Schmidt, Noah~A. Smith, and Mike Lewis. 2023.
\newblock \href {https://arxiv.org/abs/2210.03350} {Measuring and narrowing the compositionality gap in language models}.
\newblock \emph{Preprint}, arXiv:2210.03350.

\bibitem[{Rackauckas(2024)}]{ragfusion}
Zackary Rackauckas. 2024.
\newblock \href {https://doi.org/10.5121/ijnlc.2024.13103} {Rag-fusion: A new take on retrieval augmented generation}.
\newblock \emph{International Journal on Natural Language Computing}, 13(1):37–47.

\bibitem[{Raffel et~al.(2020)Raffel, Shazeer, Roberts, Lee, Narang, Matena, Zhou, Li, and Liu}]{t5}
Colin Raffel, Noam Shazeer, Adam Roberts, Katherine Lee, Sharan Narang, Michael Matena, Yanqi Zhou, Wei Li, and Peter~J Liu. 2020.
\newblock Exploring the limits of transfer learning with a unified text-to-text transformer.
\newblock \emph{Journal of machine learning research}, 21(140):1--67.

\bibitem[{Schulman et~al.(2017)Schulman, Wolski, Dhariwal, Radford, and Klimov}]{schulman2017proximal}
John Schulman, Filip Wolski, Prafulla Dhariwal, Alec Radford, and Oleg Klimov. 2017.
\newblock Proximal policy optimization algorithms.
\newblock \emph{arXiv preprint arXiv:1707.06347}.

\bibitem[{Wang et~al.(2023)Wang, Li, and Zhao}]{wang2023self}
Jinyuan Wang, Junlong Li, and Hai Zhao. 2023.
\newblock Self-prompted chain-of-thought on large language models for open-domain multi-hop reasoning.
\newblock \emph{arXiv preprint arXiv:2310.13552}.

\bibitem[{Wu et~al.(2023)Wu, Yang, Zhan, Yuan, Wong, and Chao}]{wu2023survey}
Junchao Wu, Shu Yang, Runzhe Zhan, Yulin Yuan, Derek~F Wong, and Lidia~S Chao. 2023.
\newblock A survey on llm-gernerated text detection: Necessity, methods, and future directions.
\newblock \emph{arXiv preprint arXiv:2310.14724}.

\bibitem[{Yan et~al.(2024)Yan, Gu, Zhu, and Ling}]{CRAG}
Shi-Qi Yan, Jia-Chen Gu, Yun Zhu, and Zhen-Hua Ling. 2024.
\newblock Corrective retrieval augmented generation.
\newblock \emph{arXiv preprint arXiv:2401.15884}.

\bibitem[{Yang et~al.(2023)Yang, Li, Zhang, Wang, Cheng, Li, and Xiao}]{prca}
Haoyan Yang, Zhitao Li, Yong Zhang, Jianzong Wang, Ning Cheng, Ming Li, and Jing Xiao. 2023.
\newblock \href {https://doi.org/10.18653/v1/2023.emnlp-main.326} {{PRCA}: Fitting black-box large language models for retrieval question answering via pluggable reward-driven contextual adapter}.
\newblock In \emph{Proceedings of the 2023 Conference on Empirical Methods in Natural Language Processing}, pages 5364--5375, Singapore. Association for Computational Linguistics.

\bibitem[{Yang et~al.(2018)Yang, Qi, Zhang, Bengio, Cohen, Salakhutdinov, and Manning}]{hotpotqa}
Zhilin Yang, Peng Qi, Saizheng Zhang, Yoshua Bengio, William~W Cohen, Ruslan Salakhutdinov, and Christopher~D Manning. 2018.
\newblock Hotpotqa: A dataset for diverse, explainable multi-hop question answering.
\newblock \emph{arXiv preprint arXiv:1809.09600}.

\bibitem[{Yao et~al.(2022)Yao, Zhao, Yu, Du, Shafran, Narasimhan, and Cao}]{react}
Shunyu Yao, Jeffrey Zhao, Dian Yu, Nan Du, Izhak Shafran, Karthik Narasimhan, and Yuan Cao. 2022.
\newblock React: Synergizing reasoning and acting in language models.
\newblock \emph{arXiv preprint arXiv:2210.03629}.

\bibitem[{Yu et~al.(2023)Yu, Iter, Wang, Xu, Ju, Sanyal, Zhu, Zeng, and Jiang}]{genread}
Wenhao Yu, Dan Iter, Shuohang Wang, Yichong Xu, Mingxuan Ju, Soumya Sanyal, Chenguang Zhu, Michael Zeng, and Meng Jiang. 2023.
\newblock \href {https://arxiv.org/abs/2209.10063} {Generate rather than retrieve: Large language models are strong context generators}.
\newblock \emph{Preprint}, arXiv:2209.10063.

\bibitem[{Zheng et~al.(2024)Zheng, Mishra, Chen, Cheng, Chi, Le, and Zhou}]{stepback}
Huaixiu~Steven Zheng, Swaroop Mishra, Xinyun Chen, Heng-Tze Cheng, Ed~H. Chi, Quoc~V Le, and Denny Zhou. 2024.
\newblock \href {https://arxiv.org/abs/2310.06117} {Take a step back: Evoking reasoning via abstraction in large language models}.
\newblock \emph{Preprint}, arXiv:2310.06117.

\end{thebibliography}

\appendix

\section{Case Study}
\label{sec:appendix}

To explicitly and intuitively demonstrate the effectiveness of the ERRR compared to the RRR framework, we present two examples in Table \ref{case study} comparing their rewritten queries and final outputs. In the first example, the original question is \textit{Stories USA starred which actor and comedian from "The Office"?}. The query rewriter in RRR framework produces a simplified query, \textit{actor comedian "The Office" Stories USA cast}, which merely reformulates the original question for clearer web searching. In contrast, the ERRR not only answers correctly in the  Parametric Knowledge Extraction phase but also generates refined queries such as \textit{"actor and comedian from "The Office" in Stories USA"} and \textit{"Steve Carell role in Stories USA"}, These queries not only attempt to validate the actor name of the \"The Office\" but also attempt to validate the name \textit{Steve Carell} from the parametric knowledge, enabling the retriever to source better results.

\begin{table*}[tbh]
	\centering
	\small
	\label{tab:case-study-compact}
	\begin{tabularx}{\linewidth}{@{} l >{\raggedright\arraybackslash}X @{}}
		\toprule
		\multicolumn{2}{@{}l}{\textbf{Example 1}} \\
		\midrule
		Question & Stories USA starred which actor and comedian from "The Office"? \\
		Answer   & Steven John Carell \\
		\cmidrule(r){1-2}
		\multicolumn{2}{@{}l}{\textbf{RRR}} \\
		Rewritten Query(s) & actor comedian "The Office" Stories USA cast \\
		Output & Ricky Gervais (\textit{incorrect}) \\
		\cmidrule(r){1-2}
		\multicolumn{2}{@{}l}{\textbf{ERRR}} \\
		Parametric Knowledge & The model's internal knowledge correctly identifies \textbf{Steve Carell} as the star of "Stories USA" and an actor from "The Office," providing additional (though unverified) details about his role and career. \\
		Rewritten Query(s) & actor and comedian from "The Office" in Stories USA, Steve Carell role in Stories USA \\
		Output & Steven John Carell (\textit{correct}) \\
		\midrule[\heavyrulewidth]
		\multicolumn{2}{@{}l}{\textbf{Example 2}} \\
		\midrule
		Question & What Pakistani actor and writer from Islamabad helped write for the 2012 Pakistani comedy drama sitcom, "Coke Kahani"? \\
		Answer   & Yasir Hussain \\
		\cmidrule(r){1-2}
		\multicolumn{2}{@{}l}{\textbf{RRR}} \\
		Rewritten Query(s) & Pakistani actor writer Islamabad Coke Kahani 2012 \\
		Output & Ali Abbas (\textit{incorrect}) \\
		\cmidrule(r){1-2}
		\multicolumn{2}{@{}l}{\textbf{ERRR}} \\
		Parametric Knowledge & The model incorrectly identifies \textbf{Faisal Rehman} as a writer for "Coke Kahani," but correctly notes he is a Pakistani actor and writer. This flawed parametric knowledge still provides a useful starting point for query refinement. \\
		Rewritten Query(s) & Pakistani actor and writer from Islamabad who helped write for Coke Kahani, Faisal Rehman contributions to Coke Kahani \\
		Output & Yasir Hussain (\textit{correct}) \\
    \bottomrule
	\end{tabularx}
    		
        \caption{A case study comparing the RRR and ERRR frameworks. For each example, we present the query and output from the baseline RRR, followed by the extracted parametric knowledge, refined queries, and final output from ERRR.}
	\vspace{-1.2em}
\end{table*}

In the second example, the rewritten query from RRR, \textit{Pakistani actor writer Islamabad Coke Kahani 2012}, rewrites into only a few random keywords from the original question, which fails to facilitate a high-quality search. On the other hand, the first rewritten query from ERRR, \textit{Pakistani actor and writer from Islamabad who helped write for Coke Kahani}, provides a clearer and more comprehensible query for search possibly inspired by the contextual information from the extracted parametric knowledge. The second rewritten query, \textit{Faisal Rehman contributions to Coke Kahani} aims to verify the name derived from parametric knowledge, specifically Faisal Rehman. Interestingly, even though the name is incorrect, the information retrieved subsequently clarifies that Faisal Rehman is not the correct actor and writer, which effectively rectifies the LLM's output. Together with the information gathered from the first query, this leads to a correct final answer. This example illustrates that even if the pseudo-contextual document contains inaccuracies, the ERRR framework, by concentrating on the specific needs of the LLM reader, can still retrieve the useful information, resulting in a correct outcome.

\end{document}